\pdfoutput=1

\documentclass[11pt]{article}

\usepackage[table,xcdraw]{xcolor}

\usepackage[final]{acl}
\usepackage{algorithm}
\usepackage[noend]{algpseudocode}
\usepackage{pifont}
\newcommand{\cmark}{\ding{51}}%
\newcommand{\xmark}{\ding{55}}%
\usepackage{amsmath}
\usepackage{multirow}
\usepackage{mathabx}
\usepackage{booktabs}
\usepackage{times}
\usepackage{enumitem}

\usepackage{latexsym}

\usepackage[T1]{fontenc}

\usepackage[utf8]{inputenc}

\usepackage{microtype}
\usepackage{slashbox}
\usepackage{inconsolata}

\usepackage{graphicx}
\definecolor{prunegray}{gray}{0.6}
\title{Walk and Read Less:
Improving the Efficiency of Vision-and-Language Navigation via Tuning-Free Multimodal Token Pruning}

\author{
 \textbf{Wenda Qin},
 \textbf{Andrea Burns},
 \textbf{Bryan A. Plummer},
 \textbf{Margrit Betke}
\\
Boston University
\\
 \small{
   \textbf{Correspondence:} \href{mailto:wdqin@bu.edu}{wdqin@bu.edu}
 }
}

\begin{document}
\maketitle
\begin{abstract}
Large models achieve strong performance on Vision-and-Language Navigation (VLN) tasks, but are costly to run in resource-limited environments. Token pruning offers appealing tradeoffs for efficiency with minimal performance loss by reducing model input size, but prior work overlooks VLN-specific challenges.  For example,  information loss from pruning can effectively increase computational cost due to longer walks. Thus, the inability to identify uninformative tokens undermines the supposed efficiency gains from pruning.
To address this, we propose Navigation-Aware Pruning (NAP), which uses navigation-specific traits to simplify the pruning process by pre-filtering tokens into foreground and background. For example, image views are filtered based on whether the agent can navigate in that direction. We also extract navigation-relevant instructions using a Large Language Model.  After filtering, we focus pruning on background tokens, minimizing information loss.  To further help avoid increases in navigation length, we discourage backtracking by removing low-importance navigation nodes.
Experiments on standard VLN benchmarks show NAP significantly outperforms prior work, preserving higher success rates while saving more than 50\% FLOPS\footnote{Code available: \url{https://github.com/wdqin/VLN-NAP}}.
\end{abstract}

\begin{figure}[t]
\includegraphics[width=1.0\columnwidth]{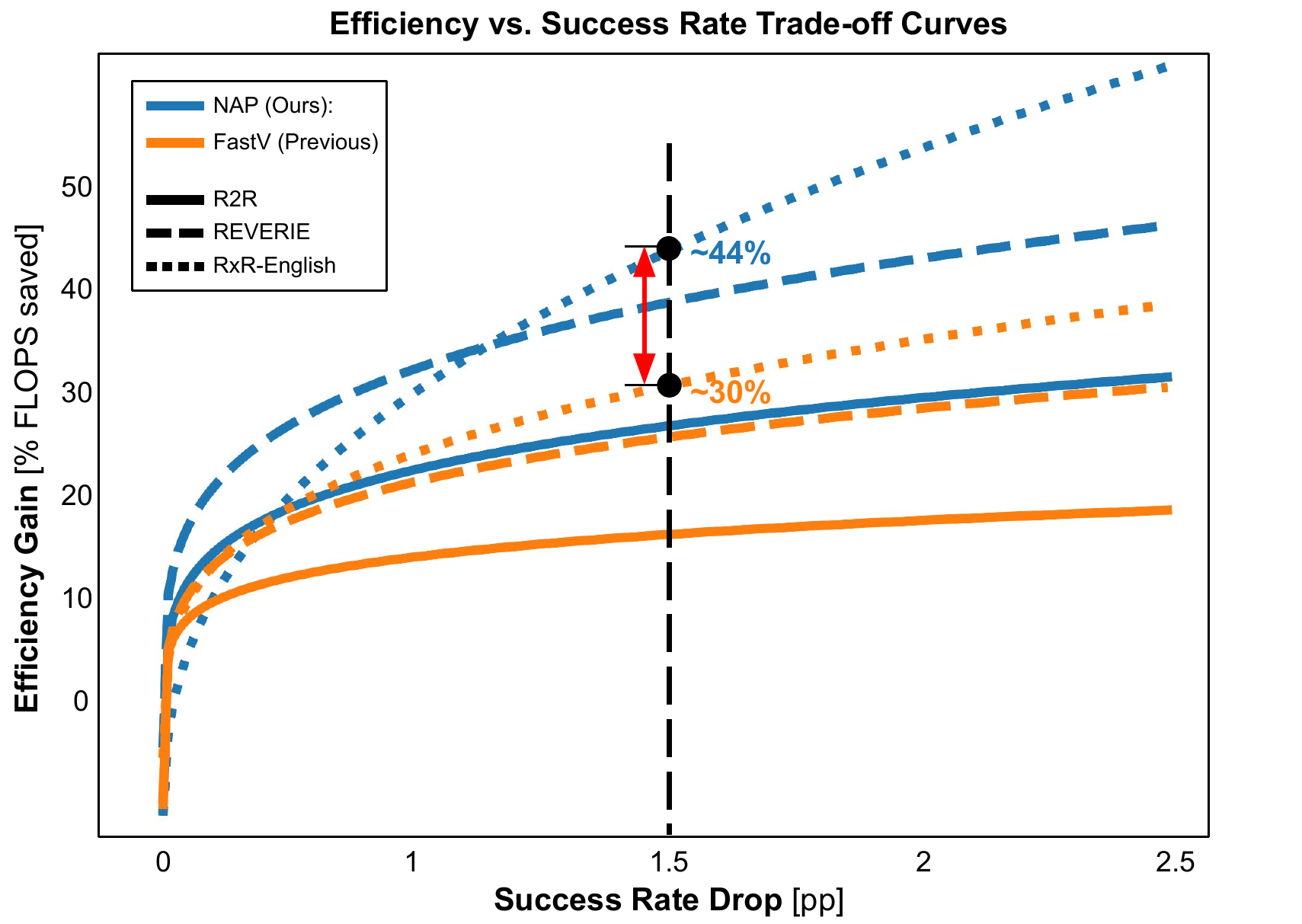}
\vspace*{-8mm}
\caption{\textbf{Efficiency-vs-Success-Rate trade-off curves} for our method NAP (blue) and FastV~\cite{chen2025image} (orange) for 3 VLN datasets. Curves are fitted based on 15 token budget settings (100\% to 30\% of original budget).
NAP consistently achieves greater efficiency for the same success rate loss compared to FastV. For example, for a 1.5\% point (pp) drop in the success rate for navigation with RxR data, NAP achieves a 14\% point gain in efficiency over FastV (red).
}
\vspace*{-3.5mm}
\label{fig:figure_1_1}
\end{figure}

\section{Introduction}
Vision-and-Language Navigation (VLN) 
evaluates an AI agent's ability to navigate an environment following a natural language instruction \cite{fried2018speaker, tan2019learning, hong2021vln, chen2021history, chen2022think, wang2024magic}. However, sometimes the computational costs for high-performing models are too high for hardware-limited agents, 
underscoring the need to prioritize  {\bf navigation efficiency}. Token pruning improves computational efficiency by reducing input size, offering a trade-off between performance and cost. For example, as shown in Fig.~\ref{fig:figure_1_1},  methods like FastV~\cite{chen2025image} can reduce the number of floating point operations per second (FLOPS) by 
30\%
while only suffering from a 1.5\% drop in navigation 
success rate 
(SR) on 
RxR-English 
 \cite{ku2020room}. Additionally, tuning-free token pruning allows for the same model to flexibly adapt to different hardware constraints by adjusting pruning rates without costly retraining.

\begin{figure}[t!]
\includegraphics[width=0.9\columnwidth]{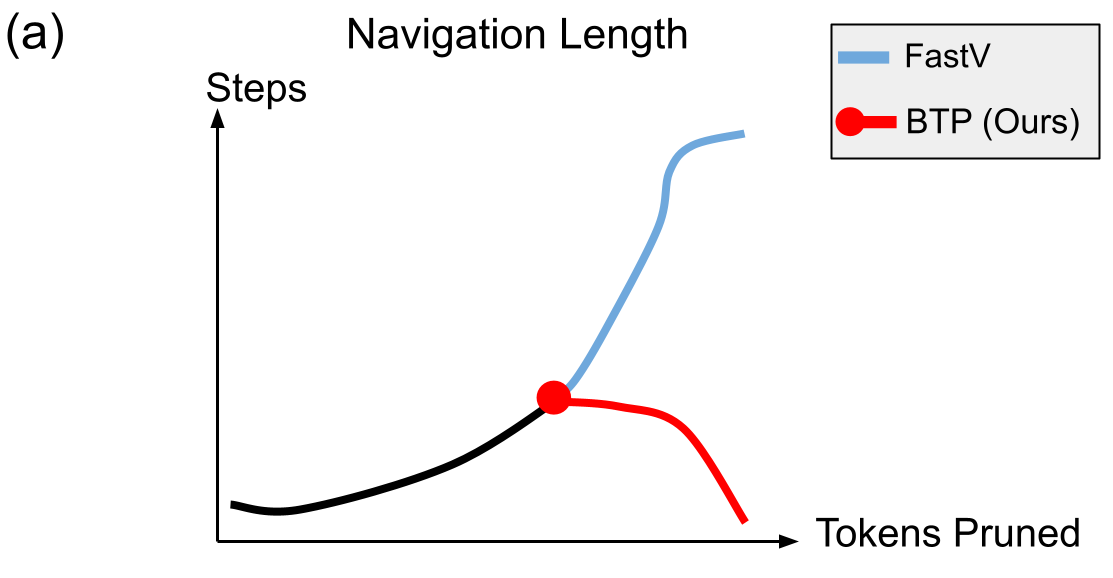}
\includegraphics[width=\columnwidth]{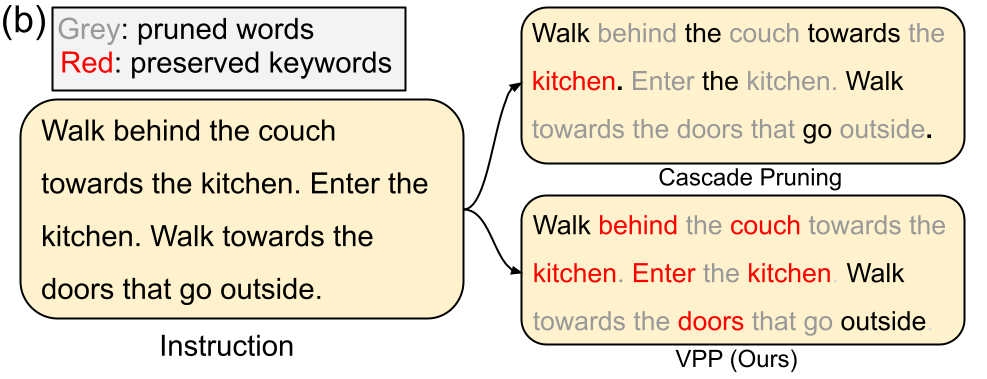}
\vspace*{-9mm}
\caption{(a) Navigation length vs. number of pruned tokens. As token pruning strategies, BTP reduces navigation length while FastV increases it. (b) Example of instruction pruning. By prioritizing navigation-irrelevant words, VPP preserves more useful information in the instruction than Cascade pruning~\cite{wang2021spatten} with the same token budget.}
\vspace*{-3.5mm}
\label{fig:figure_1_2}
\end{figure}

A drawback of existing token pruning strategies~\cite{bolya2022token, zhang2024sparsevlm, chen2025image} is that they are designed for general Vision-and-Language Models (VLMs), ignoring the temporal dependence inherent in VLN tasks. 
This limits their ability to reducing navigation computation cost as pruned tokens may contain useful context to aids the agent’s decisions. Without them, the agent becomes less certain and often backtracks to unvisited nodes for additional clues.  This results in longer paths and additional computation (see blue curve in Fig.~\ref{fig:figure_1_2}(a)), ultimately reducing the efficiency gains of pruning. In effect, the widely used attention-based textual pruning~\cite{goyal2020power,wang2021spatten} often fails to distinguish relevant from irrelevant instruction tokens. We find VLN models tend to assign high attention scores to punctuation and function words (e.g., ``,'' and ``the''), as shown in Fig.~\ref{fig:figure_1_2}(b) and supported by prior work~\cite{clark2019does}. As a result, important content tokens may be pruned instead, leaving instructions uninformative and impairing the agent’s ability to navigate effectively.

\begin{figure*}[t]
\centering
\includegraphics[width=\textwidth]
{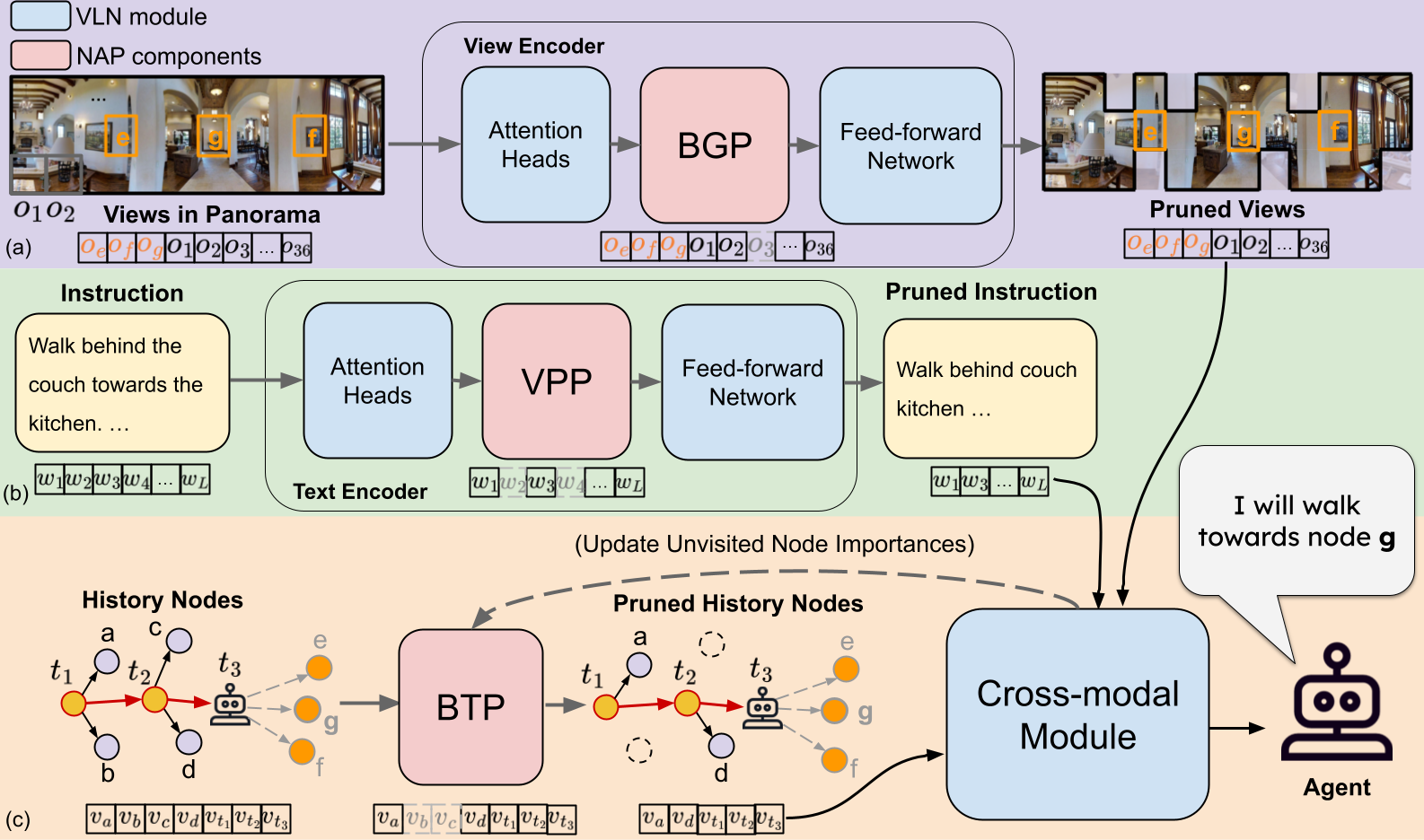}
\vspace*{-8mm}
  \caption{An overview of NAP components (BGP, VPP, BTP) on a VLN model. (a) BGP reduces the size of visual inputs by pruning the non-critical background views from $\{o_1,o_2,...,o_{36}\}$ (details in Sec.~\ref{sec:bgp}). (b) VPP prunes the irrelevant word tokens from the instruction tokens $\{w_1,w_2,...,w_L\}$ to shorten textual inputs (described in Sec.~\ref{subsec:VPP}), and (c) BTP prunes the unvisited history nodes from $\{v_a,v_b,...,v_h\}$ on the navigation map constructed by the VLN models to discourage backtracking (discussed in Sec.~\ref{subsec:BTP}).}
  \label{fig:figure_2}
\end{figure*}

To address these challenges, we propose \textbf{N}avigation-\textbf{A}ware \textbf{P}runing (NAP), a framework tailored for navigation tasks that enhances navigation efficiency by shortening the navigation duration (``walk less''), and prioritizing pruning navigation-irrelevant tokens  (``read less''), achieving a significantly improved SR–FLOPS trade-off compared to prior methods (Fig.~\ref{fig:figure_1_1}).
NAP begins by separating views in the input panorama into action views (e.g., views corresponding to \textit{e}, \textit{f}, \textit{g} in Fig.~\ref{fig:figure_2} (a)) and background views ($o_1, o_2$ in Fig.~\ref{fig:figure_2} (a)). We discovered that background views provide contextual information highly amenable to pruning, compared to action views. Our results demonstrate that selectively removing background tokens achieves substantial computational reduction with a minimal loss in SR. This process forms the basis of \textbf{B}ack\textbf{G}round \textbf{P}runing (BGP) in NAP.

We find that selectively pruning unvisited nodes (e.g., b, c in Fig.~\ref{fig:figure_2}(c)) reduces path length (red in Fig.~\ref{fig:figure_1_2}(a)), while preserving backtracking benefits when the number of retained nodes is properly tuned. To enable this, NAP introduces BackTracking Pruning (BTP), which removes unvisited nodes with low-importance scores from previous steps. 

To enable the model to ``read less,'' we introduce \textbf{V}ocabulary \textbf{P}riority \textbf{P}runing (VPP) to distinguish and prune uninformative instruction tokens as part of NAP. VPP constructs a \textit{vocabulary of irrelevance} by prompting a large language model (e.g., LLaMA 3~\cite{dubey2024llama}) to identify terms that are non-essential for navigation. This allows VPP to prioritize pruning irrelevant tokens while preserving both LLM-identified and potentially important unseen words. As shown in Fig.~\ref{fig:figure_1_2} (b), important words like “couch,” “enter,” and “doors” are successfully retained by VPP, whereas the prior method \cite{wang2021spatten} fails to do so.

\noindent
To summarize, our contributions are as follows:

\begin{itemize}[nosep,leftmargin=*]
    \item  We propose Navigation-Aware Pruning (NAP), which includes Background Pruning (BGP) to reduce visual inputs with minimal success rate loss, and BackTracking Pruning (BTP) to improve path efficiency by limiting backtracking.
    \item NAP introduces Vocabulary Priority Pruning (VPP), which leverages an LLM  to identify and prune navigation-irrelevant instruction tokens, thus reducing input size. 
    \item By combining BGP, BTP, and VPP, NAP achieves up to 2$\times$ speedups over the original model and 1.25$\times$ over the state-of-the-art pruning baseline with a smaller success rate drop (2-3~pp). These improvements are observed across multiple VLN models and datasets for both discrete and continuous environments.
\end{itemize}






\section{Related Work}

\textbf{VLN Datasets and Models}. Multiple datasets \citep{anderson2018vision, ku2020room, qi2020reverie, zhu2021soon, hong2022bridging} have been curated to address VLN challenges that include long paths~\citep{ku2020room}, object localization~\citep{qi2020reverie,zhu2021soon}, and continous environment~\citep{krantz2020beyond}. To 
navigate through the environment, agents rely on transformer-based architectures 
for action prediction 
\citep{chen2021history, chen2022think, an2023bevbert, zhao2022target, huo2023geovln, wang2023dual, li2023panogen, wang2023gridmm, qiao2023hop+, li2023improving, wang2024vision}. 

Advancements include history modules for navigation memory \citep{chen2021history}, backtracking mechanisms with topological mapping \citep{chen2022think}, enabling BTP as an efficiency-performance trade-off, and the use of synthesized visual data \citep{li2023panogen}. \citet{wang2024vision} further improved navigation accuracy with a debiasing strategy.
Regarding VLN efficiency, 
\citet{Qiao_2023_ICCV} explored Parameter-Efficient Transfer Learning, while 
\citet{zhu2024minivln} and 
\citet{wang2024magic} applied knowledge distillation to train smaller VLN models. These methods require new modules or model reconstruction, while our approach does not.

\noindent
\textbf{Token Reduction (Pruning and Merging)}. 
Recent works \cite{goyal2020power, wang2021spatten, kim2022learned, liang2022not, bolya2022token, wei2023joint, wang2024zero, zhang2024sparsevlm, chen2025image} primarily rely on attention scores to reduce sequence lengths in transformers. PoWER-BERT~\cite{goyal2020power} prunes input tokens based on self-attention. SpAtten~\cite{wang2021spatten} 
formally introduced 
cascade token and head pruning. 
\citet{liang2022not} addressed information loss by fusing pruned tokens, while 
\citet{bolya2022token} focused on removing feature redundancy. 
\citet{chen2025image} prunes all unimportant tokens at a specific layer of the VLM. SparseVLM \cite{zhang2024sparsevlm} leverages external LLMs to assess visual token importance.  Previous strategies assume attention scores reflect token importance, failing under high text pruning rates. They also ignore the impact of view token pruning on 
efficiency. We are the first to propose VLN-specific
pruning to address these challenges. 


\section{The NAP Method: BGP, BTP, and VPP}



In VLN, an agent navigates based on an instruction of words \(I = \{w_1, \dots, w_L\}\). At each step, it processes views from a panorama \(P = \{o_1, \dots, o_N\}\)~\cite{fried2018speaker} and selects navigable locations \(a \in A\) until it predicts a “STOP” action \(a_{\text{stop}}\). Using the Matterport 3D simulator \cite{anderson2018vision,ku2020room,qi2020reverie}, navigation is successful if the agent stops within 3 meters of the target; an agent in REVERIE~\cite{qi2020reverie} must also locate the target object.

Most VLN models split the views in panoramas $P = \{o_{1},...,o_{N}\}$ into action views $O_{act} = \{o_1,...,o_n\}$ and background views $O_{bgr} = \{o_{n+1},...,o_{N}\}$ providing visual context. The action views $\{o_1,...,o_n\}$ correspond to navigable nodes $\{a_1,...,a_n\}$ indicating navigable locations at the current step.
In addition to views, DUET-based~\cite{chen2022think} models~\cite{li2023panogen,wang2024vision,wang2024magic,zhu2024minivln} adopt a topological map as ``history'' to track the unvisited nodes
$V_{unvisited} = \{v_1, v_2, ..., v_m\}$ and visited nodes $V_{visited} = \{v_{m+1}, v_{m+2}, ..., v_M\}$, with $V_{unvisited}$ enabling backtracking actions $\{a^{bt}_{1},...,a^{bt}_{m}\}$. The model’s inputs are thus tri-modal, including $I$, $P$, and $V$.
At each step, the model selects an action from stop, navigate, or backtrack using the policy \(a = \pi(I,P,V)\), where \(a \in \{a_{\text{stop}}, a_1,\dots,a_n\} \cup \{a^{bt}_1,\dots,a^{bt}_m\}\).

Given an instruction $I = \{w_1, w_2, ..., w_L\}$, views $P = \{o_1,...,o_{N}\}$, and history nodes $V = \{v_1, v_2, ..., v_M\}$, token pruning reduces sequence lengths with a certain process, denoted as function $f$. 
We obtain the pruned sequences $I' = f_{I}(w_1,...,w_L)$, $P' = f_{P}(o_1,...,o_N)$, and $V' = f_{V}(v_1,...,w_M)$, such that $|I'| < |I|$, $|P'| < |P|$, $|V'| < |V|$. The VLN model then selects an action \(a = \pi(I',P',V')\) with lower computational cost. 

Efficient token pruning hinges on accurately assessing token importance.
Our NAP framework for this problem includes three components: BackGround Pruning (BGP) for views (Sec.~\ref{sec:bgp}), BackTracking Pruning (BTP) of unvisited nodes (Sec.~\ref{subsec:BTP}), and Vocabulary Priority Pruning (VPP) for instruction tokens (Sec.~\ref{subsec:VPP}), which are integrated into NAP within VLN models (e.g., GOAT~\cite{wang2024vision}) as shown in Fig.~\ref{fig:figure_2}.

\subsection{Background Pruning (BGP)}
\label{sec:bgp}

In this section we discuss how we boost efficiency by pruning visual tokens.
During the navigation process, panoramic views $P$ are projected into a feature space using a pre-trained feature extractor, such as CLIP-B/16 \cite{radford2021learning}, and enhanced with learned positional encodings representing the viewing direction. This results in a sequence of $N$ view tokens, each associated with a feature $x$. For simplicity, we denote the encoded view token sequence as $P$ and each view feature token as $o$.
Our token importance scores come from the transformer blocks processing $P$.
Specifically, each transformer block consists of a self-attention module with attention heads, followed by a feed-forward network. At each head $h$ in layer $i$, self-attention scores are computed to quantify the dependencies between tokens:
\begin{equation}
\text{Attn}_{h_i} = \text{softmax}[Q_i \times K_i^T ],
\end{equation} 
where $Q_i = P_i \times W^{h_i}_{q}$, $K_i = (P_i \times W^{h_i}_{k})$, and $W^{h_i}_{q}$, $W^{h_i}_{k}$ are trained weight matrices.
The resulting $\text{Attn}_{h_i}$ is a $N \times N$ matrix, where each entry $\text{Attn}_{h_i}[o_m,o_n]$ represents the normalized dependency of view $o_n$ on view $o_m$. A higher value of $\text{Attn}_{h_i}[o_m,o_n]$ usually indicates that view $o_n$ has a stronger influence on constructing the updated representation of $o_m$ in the output $P_{i+1}$.

To determine the overall importance of each token, we aggregate the attention scores by summing each column of $\text{Attn}_{h_i}$ across all attention heads $H_i$ at layer $i$. This gives a vector of scores indicating the influence of each token $o'$ on all tokens:
\begin{equation}
    \text{Score}(o') = \sum_{h_i \in H_i}\sum_{o \in P}\text{Attn}_{h_i}[o,o'].
    \label{eq:score}
    \end{equation}
We interpret $\text{Score}(o')$ as the \textbf{importance score} of view $o'$, which guides the pruning process by identifying less influential background views. Similar scores can be obtained for history nodes $v$ (described in Sec.~\ref{subsec:BTP}) and words $w$ (discussed in Sec.~\ref{subsec:VPP}) from the cross-modal module and language module, respectively.
The module concludes by computing the latent feature 
$Z_{i} = \text{concatenate}(A_{h_i} \times (P_i \times W^{h_i}_{v}))$, where the concatenation is performed across all heads and $W^{h_i}_{v}$ is the weight matrix for each head. 

\begin{figure}[t]
\centering
\includegraphics[width=\columnwidth]{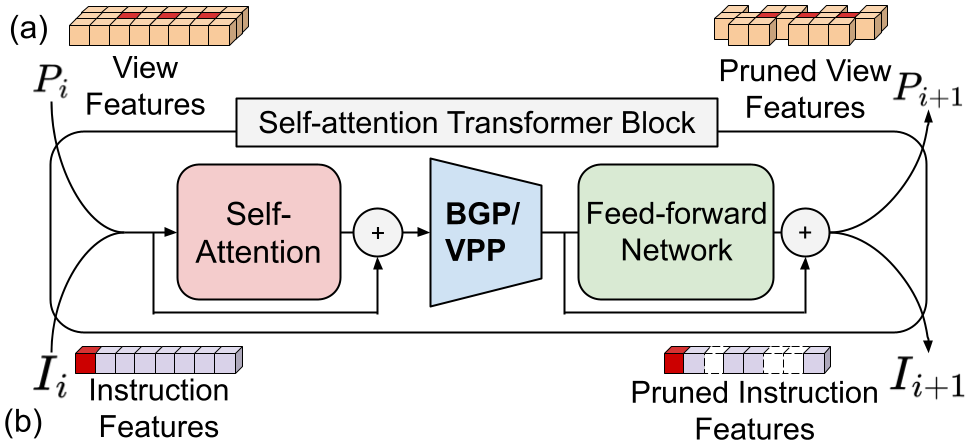}
\vspace*{-8mm}
\caption{BGP (a) and VPP (b) processes. Both BGP and VPP prune view feature tokens based on their importance score after the self-attention module and the residual block at each transformer layer in a VLN model.}
\label{fig:bgp_vpp}
\end{figure}

BGP reduces the size of \(O_{bgr}\) in \(P\) by removing \(k_{\text{BGP}}\) tokens at each layer of the vision transformer encoder (see Fig.~\ref{fig:bgp_vpp}). Following \citet{bolya2022token}, BGP is applied after computing the attention matrix \(\text{Attn}_{H_i}\) for \(P\) and before passing \(Z_i\) to the residual connection and feed-forward network. BGP retains all action view tokens \(O_{act}\) while pruning the \(k_{\text{BGP}}\) tokens with the lowest \(\text{Score}(o)\) from \(O_{bgr}\) (i.e., removing their \(o\) and \(z\) values from $O_{bgr}$ and \(Z_i\)).  Action views are given, and relate to the views where the agent can perform an action.  For example, if the agent can move to the right, then the view in that direction is an action view. The pruned \(P_i\) and \(Z_i\) are then processed to yield \(P_{i+1}\). Consequently, the final output \(P_{L+1}\) contains \(k_{\text{BGP}} \cdot L\) fewer tokens than the original \(P\), resulting in a smaller visual input for action prediction.

This procedure can be thought of as separating tokens into foreground (action) and background (context), and then only pruning background tokens.  In effect, we leverage \emph{a priori} knowledge to simplify our token selection problem.  This helps minimize the possibility that we remove key tokens that would result in longer navigation paths, which is further reduced by BTP in the next section.

\begin{figure}[t]
\centering
\includegraphics[width=\columnwidth]{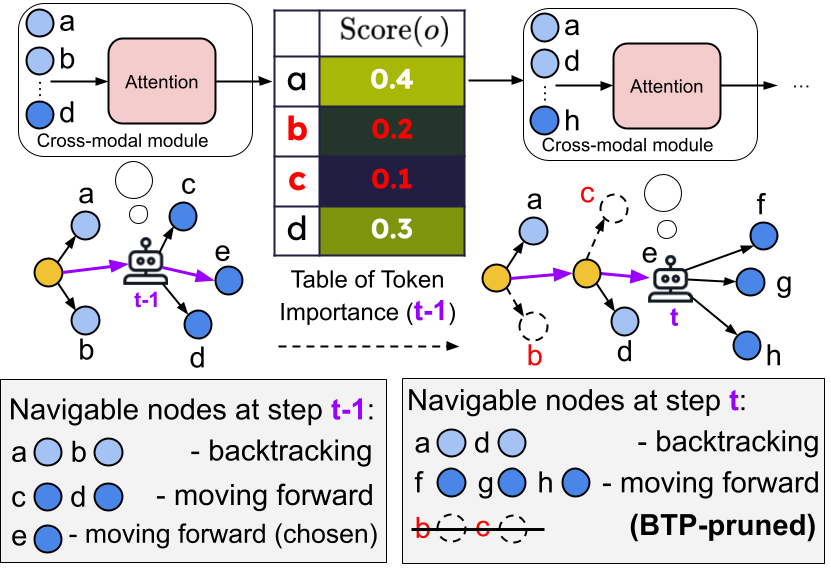}
\vspace*{-8mm}
\caption{The BTP 
process. By keeping the number of backtracking nodes in the history less or equal to the threshold $k_{BTP}=2$, (nodes a, d), and removing the rest (nodes b, c), the agent has fewer backtrack choices, thus is more likely to move forward (f, g, h) or stop.}
\label{fig:btp}
\end{figure}


\subsection{Backtracking Pruning (BTP)}
\label{subsec:BTP}
Backtracking in DUET-based VLN models~\cite{chen2022think} refers to returning to unvisited nodes from previous navigation steps.  However, as discussed earlier, pruning tokens as done in Sec.~\ref{sec:bgp} and Sec.~\ref{subsec:VPP}) can increase navigation length.  To address this, we propose BTP, which removes a subset of inconsequential unvisited nodes $V_{unvisited}$ from the history input $V$ (nodes b \& c in Fig.~\ref{fig:btp}) to discourage backtracking. Specifically, $V_{unvisited} = \{v_{1}, v_{2}, ..., v_m\}$ contains action view tokens from earlier steps that were not selected, i.e., $\{v_{1},v_{2},...,v_{m}\} = \{o_{1},o_{2},...,o_{m}\}$, where
\{$o_{1},o_{2}, ...,o_{m}\} \in O_{act}^{t'}$ at previous steps $t'<t$. To determine the most crucial nodes for successful navigation, we track the importance scores of these tokens from the attention heads of the last cross-modal transformer block at each step (e.g., 0.4 for node $a$ in Fig.~\ref{fig:btp}). At the beginning of the next step, BTP retains the top $k_{\text{BTP}}$ unvisited nodes based on their latest $\text{Score}(o)$ and discards the remainder, so that $V_{unvisited} = \{v_1,...,v_{k_\text{BTP}}\}$.

BTP offers two main benefits: It reduces the size of the history input \(V\) to the cross-modal module, lowering computational cost, and it limits backtracking options from \(\{a'_1, \dots, a'_m\}\) to \(\{a'_1, \dots, a'_{k_\text{BTP}}\}\), enhancing path efficiency by reducing potential unnecessary backtracking.

\subsection{Vocabulary Priority Pruning (VPP)}
\label{subsec:VPP}
\begin{figure}[t]
\centering
\includegraphics[width=\columnwidth]{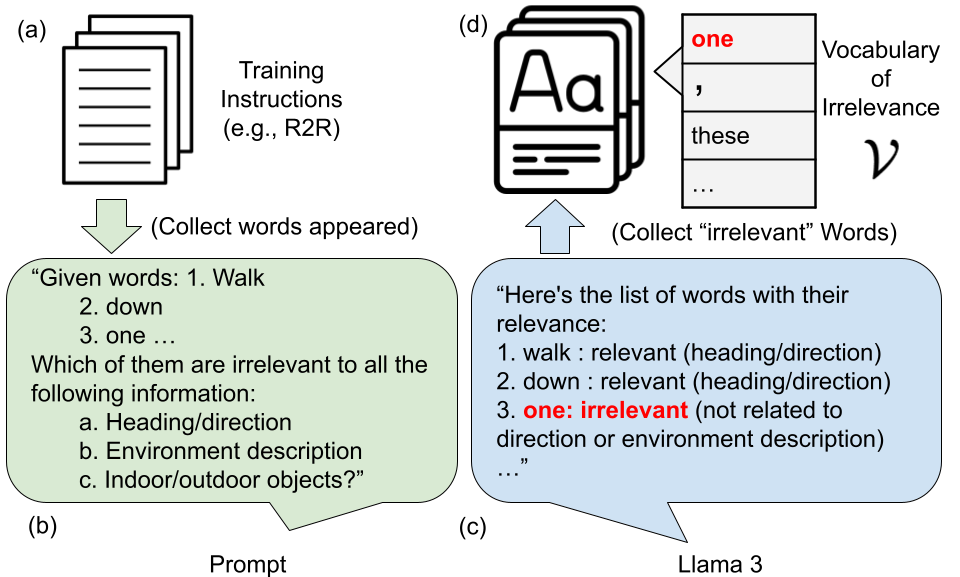}
\vspace*{-7mm}
\caption{The construction of the ``vocabulary of irrelavance'' $\mathcal{V}$, as a 4-step process, (a) collecting lexicon of words appeared in the training instructions, (b) prompting a LLM with words in the lexicon, (c) Classifying relevant and irrelevant words by LLM, and (d) grouping words irrelevant into the vocabulary. The resulted vocabulary helps identify words likely to waste computation, and thus should be pruned.}
\label{fig:vpp}
\end{figure}

VPP operates similarly to BGP from  Sec.~\ref{sec:bgp} by pruning instruction tokens $w \in I$ based on the importance $\text{Score}(w)$ derived from the attention mechanism of the language transformer block (see Fig.~\ref{fig:bgp_vpp}). However, we found that the attention scores are a noisy measure of importance. For example, as shown in Table~\ref{tab:vpp_example}, non-informative tokens, such as punctuation, often receive a high $\text{Score}(w)$ and are retained, thereby wasting the token budget. Thus, as we did for BGP, we simplify the token selection problem via filtering using task-specific knowledge.  However, rather than using action constraints, we prompt an LLM to construct a ``vocabulary of irrelevance'' $\mathcal{V}$ that identifies unimportant words based on its general knowledge rather than attention scores (illustrated in Fig.~\ref{fig:vpp}).
We tokenize the training data to form a lexicon and prompt Llama 3~\cite{dubey2024llama} to label each word as “relevant” or “irrelevant” for navigation based on its association with direction/heading, environment description, or indoor/outdoor objects. Words labeled “irrelevant” were compiled into $\mathcal{V}$; see Appendix~\ref{app:vocab_exam} for an example prompt and vocabulary.

At each layer of the language transformer encoder, VPP prunes $k_{\text{vpp}}$ instruction tokens via a two-step process. First, it filters out all words present in $\mathcal{V}$. If the number of filtered tokens is less than $k_{\text{vpp}}$, the remaining tokens are pruned based on their attention scores $\text{Score}(w)$ until the requirement is met. Otherwise, if more tokens are filtered than necessary, some are reinstated according to $\text{Score}(w)$ to preserve additional information. The VPP process is detailed in Algorithm.~\ref{alg:vpp}.

\begin{algorithm}[t]
\small
    \caption{Vocabulary Priority Pruning}
    \label{alg:vpp}
    \begin{algorithmic}[1] 
    \Procedure{VPP}{$I,\mathcal{V}, \text{Scores}, k_{\text{vpp}}$}
    \State $I_{r}$ $\gets$ empty sequence \Comment{retained tokens}
    \State $I_{p}$ $\gets$ empty sequence
    \Comment{pruned tokens}
    \State Sort $w$'s in $I$ by $\text{Score}(w)$ from high to low
\For{$w$ in $I$}
    \If{$w$ not in $\mathcal{V}$} 
        \State $I_r$.add($w$)
    \Else
        \State $I_p$.add($w$)
    \EndIf
\EndFor
\If {$k_{\text{vpp}} >$  length($I$) - length($I_{r}$)}  
    \For{$i = 1$ to $k_{\text{vpp}}$}
        \State $I_r$.add($w_i$)
    \EndFor
\Else
    \For {$w$ in $I_{p}$}
        
        \If {length($I$) - length($I_{r}$) = $k_{\text{vpp}}$}
            \State \textbf{break}
        \EndIf
        \State $I_{r}$.add($w$)
    \EndFor
\EndIf
\State \textbf{return} $I_{r}$
            
    \EndProcedure
    \end{algorithmic}
\end{algorithm}

VPP more effectively retains essential words in instructions, enabling the agent to function normally even at higher pruning rates, than attention-based strategies (Table~\ref{tab:vpp_example}). Importantly, the vocabulary is constructed \textbf{prior} to navigation (\textbf{offline}) and, thus, \textbf{no} LLM computation overhead is introduced during the actual navigation process.


\section{Experiments}

\begin{table}[t]
\small
\setlength{\tabcolsep}{1pt}
\begin{tabular}{c|l|l}
\hline
\multicolumn{1}{c|}{\begin{tabular}[c]{@{}c@{}}Retain \\ \% \end{tabular}} & \multicolumn{1}{c|}{Method} & \multicolumn{1}{c}{\begin{tabular}[c]{@{}c@{}}Instruction Tokens \\ (grey ones are pruned)\end{tabular}}                                                                                          \\ \hline
100                                                                                       & -                                   & \begin{tabular}[c]{@{}l@{}}\textless{}s\textgreater Exit the room \textbf{.} Turn right \textbf{.} Start down\\  the stairs and stop 3 steps down \textbf{.} \textless{}/s\textgreater{}\end{tabular}                        \\
50                                                                                        & VPP                                 & {\begin{tabular}[c]{@{}l@{}}\textless{}s\textgreater Exit \textcolor{prunegray}{the} room \textcolor{prunegray}{\textbf{.}} Turn right \textcolor{prunegray}{\textbf{.}} Start \textcolor{prunegray}{down}\\  \textcolor{prunegray}{the} stairs \textcolor{prunegray}{and} stop \textcolor{prunegray}{3} steps \textcolor{prunegray}{down \textbf{.}}  \textcolor{prunegray}{\textless{}/s\textgreater{}}\end{tabular}} \\
50                                                                                        & Att.\ Scores                    & \begin{tabular}[c]{@{}l@{}}\textless{}s\textgreater \textcolor{prunegray}{Exit} the \textcolor{prunegray}{room} \textbf{.} \textcolor{prunegray}{Turn right} \textbf{.} Start \textcolor{prunegray}{down}\\  \textcolor{prunegray}{the} stairs \textcolor{prunegray}{and stop} 3 \textcolor{prunegray}{steps down} \textbf{.} \textcolor{prunegray}{\textless{}/s\textgreater{}}\end{tabular}                        \\ \hline
25                                                                                        & VPP                                 & \begin{tabular}[c]   {@{}l@{}}\textless{}s\textgreater Exit \textcolor{prunegray}{the room \textbf{.}} \textcolor{prunegray}{Turn} right \textcolor{prunegray}{\textbf{.}} Start \textcolor{prunegray}{down}\\  \textcolor{prunegray}{the stairs and stop 3} steps \textcolor{prunegray}{down \textbf{.}} \textcolor{prunegray}{\textless{}/s\textgreater{}}\end{tabular}                   \\
25                                                                                        & Att.\ Scores                    & \begin{tabular}[c]
{@{}l@{}}\textless{}s\textgreater \textcolor{prunegray}{Exit} the \textcolor{prunegray}{room \textbf{.} Turn right} \textbf{.} \textcolor{prunegray}{Start down}\\  \textcolor{prunegray}{the stairs and stop} 3 \textcolor{prunegray}{steps down} \textbf{.} \textcolor{prunegray}{\textless{}/s\textgreater{}}\end{tabular}   
                       \\ \hline
\end{tabular}
\vspace*{-3mm}
\caption{Instruction tokens retained under different pruning rates. VPP is more effective in preserving key information 
than attention-score pruning (Cascade).}
\label{tab:vpp_example}
\end{table}

\begin{table*}[t!]
\small
\setlength{\tabcolsep}{2pt}
\hspace*{-2mm}
\begin{tabular}{lcccccccccccc}
\hline
\multicolumn{1}{c|}{}                         & \multicolumn{4}{c|}{\textbf{R2R}}                                                                                    & \multicolumn{4}{c|}{\textbf{RxR-English}}                                                                            & \multicolumn{4}{c}{\textbf{REVERIE}}                                                            \\ \cline{2-13} 
\multicolumn{1}{c|}{\multirow{-2}{*}{Method}} & \multicolumn{1}{c|}{Seen$\uparrow$} & \multicolumn{1}{c|}{Unseen$\uparrow$} & \multicolumn{1}{c|}{Steps$\downarrow$} & \multicolumn{1}{c|}{FLOPS\%$\downarrow$} & \multicolumn{1}{c|}{Seen$\uparrow$} & \multicolumn{1}{c|}{Unseen$\uparrow$} & \multicolumn{1}{c|}{Steps$\downarrow$} & \multicolumn{1}{c|}{FLOPS\%$\downarrow$} & \multicolumn{1}{c|}{Seen$\uparrow$} & \multicolumn{1}{c|}{Unseen$\uparrow$} & \multicolumn{1}{c|}{Steps$\downarrow$} & FLOPS\%$\downarrow$ \\ \hline
\multicolumn{13}{c}{\cellcolor[HTML]{EFEFEF}Upper Bound, 100\% Tokens}                                                                                                                                                                                  \\
\multicolumn{1}{l|}{VLN-GOAT}                & 84.8                                & 78.1                                  & 7.3                                    & \multicolumn{1}{c|}{100}                & 75.2                                & 69.6                                  & 8.1                                    & \multicolumn{1}{c|}{100}                & 80.7                                & 53.8                                  & 10.1                                   & 100                \\ \hline 

\multicolumn{13}{c}{\cellcolor[HTML]{EFEFEF} Retain $70\% \pm 2\%$ Tokens
}    
\\
\multicolumn{1}{l|}{Random}                  & 80.4                                & 74.1                                  & 6.8                                    & \multicolumn{1}{c|}{76.1}               & 69.2                                & 64.5                                  & 8.4                                    & \multicolumn{1}{c|}{68.6}             & 72.2                                & 49.0                                  & 11.1                                   & 82.1             \\
\multicolumn{1}{l|}{FastV$^1$}               & 81.7                                & 74.9                                  & 6.9                                    & \multicolumn{1}{c|}{78.3}               & 73.2                                & 68.1                                  & 8.4                                    & \multicolumn{1}{c|}{70.7}             & 78.7                       & 51.1                         & 10.3                                   & 76.9             \\
\multicolumn{1}{l|}{Cascade Pruning$^2$}     & 81.2                                & 75.3                                  & 6.9                                    & \multicolumn{1}{c|}{76.7}               & 73.3                                & 68.5                                  & 8.4                                    & \multicolumn{1}{c|}{68.6}             & 77.0                                & 52.1                                  & 10.5                                   & 76.8             \\
\multicolumn{1}{l|}{Cascade + ToMe$^3$}      & \textbf{82.1}                       & \textbf{75.6}                         & 6.7                                    & \multicolumn{1}{c|}{74.4}               & 73.3                                & 68.4                                  & 8.3                                    & \multicolumn{1}{c|}{69.7}             & 77.5                                & 51.9                                  & 10.4                                   & 76.4             \\
\multicolumn{1}{l|}{\textbf{NAP (Ours)}}           & 81.8                                & 74.5                                  & \textbf{6.5}                           & \multicolumn{1}{c|}{\textbf{68.3}}      & \textbf{73.5}                       & 68.4                                  & 8.4                                    & \multicolumn{1}{c|}{\textbf{65.3}}    & \textbf{78.9}                                & \textbf{53.1}                                  & \textbf{10.2}                                   & \textbf{71.0}    \\ \hline
\multicolumn{13}{c}{\cellcolor[HTML]{EFEFEF} Retain $60\% \pm 2\%$ Tokens
}                                                                                                                                                                                                                                                                                      \\
\multicolumn{1}{l|}{Random}                  & 74.2                                & 68.2                                  & 8.2                                    & \multicolumn{1}{c|}{70.4}               & 65.5                                & 59.9                                  & 8.5                                    & \multicolumn{1}{c|}{59.7}             & 71.7                                & 47.3                                  & 10.8                                   & 66.6             \\
\multicolumn{1}{l|}{FastV$^1$}               & 78.7                                & 68.8                                  & 8.0                                    & \multicolumn{1}{c|}{71.6}               & 70.9                                & 65.8                                  & 8.5                                    & \multicolumn{1}{c|}{61.8}             & 77.0                                & 51.0                                  & 10.3                                   & 65.8             \\
\multicolumn{1}{l|}{Cascade Pruning$^2$}     & 78.0                                & 72.1                                  & 7.7                                    & \multicolumn{1}{c|}{65.8}               & 72.1                                & 66.9                                  & 8.6                                    & \multicolumn{1}{c|}{59.7}             & 77.2                                & 50.7                                  & 10.3                                   & 63.6             \\
\multicolumn{1}{l|}{Cascade + ToMe$^3$}      & 78.2                                & 71.6                                  & 7.4                                    & \multicolumn{1}{c|}{66.9}               & 72.2                                & 66.8                                  & 8.5                                    & \multicolumn{1}{c|}{61.0}             & 76.7                                & 51.0                                  & 10.4                                   & 65.6             \\
\multicolumn{1}{l|}{\textbf{NAP (Ours)}}           & \textbf{82.6}                       & \textbf{75.2}                         & \textbf{7.0}                           & \multicolumn{1}{c|}{\textbf{60.0}}      & \textbf{73.3}                       & \textbf{68.0}                         & 8.5                           & \multicolumn{1}{c|}{\textbf{56.2}}    & \textbf{79.1}                       & \textbf{52.2}                         & \textbf{10.1}                          & \textbf{58.9}    \\  \hline
\multicolumn{13}{c}{\cellcolor[HTML]{EFEFEF} Retain $50\% \pm 2\%$ Tokens
}                                                                                                                                                                                                                                                                                          \\
\multicolumn{1}{l|}{Random}                  & 69.3                                & 65.4                                  & 10.0                                   & \multicolumn{1}{c|}{71.6}               & 59.3                                & 54.7                                  & \textbf{8.6}                                    & \multicolumn{1}{c|}{51.5}             & 71.1                                & 44.9                                  & 10.8                                   & 53.0             \\
\multicolumn{1}{l|}{FastV$^1$}               & 74.3                                & 67.0                                  & 9.8                                    & \multicolumn{1}{c|}{75.6}               & 64.8                                & 60.4                                  & 8.7                                    & \multicolumn{1}{c|}{54.3}             & 75.6                                & 47.9                                  & 10.6                                   & 55.9             \\
\multicolumn{1}{l|}{Cascade Pruning$^2$}     & 74.1                                & 70.7                                  & 9.8                                    & \multicolumn{1}{c|}{70.0}               & 68.3                                & 63.6                                  & 8.8                                    & \multicolumn{1}{c|}{52.9}             & 75.2                                & 46.3                                  & 10.5                                   & 51.4             \\
\multicolumn{1}{l|}{Cascade + ToMe$^3$}      & 77.4                                & 71.1                                  & 7.6                                    & \multicolumn{1}{c|}{62.6}               & 68.3                                & 63.7                                  & 8.7                                    & \multicolumn{1}{c|}{53.8}             & 63.3                                & 42.8                                & 15.2                                   & 75.7             \\
\multicolumn{1}{l|}{\textbf{NAP (Ours)}}           & \textbf{81.4}                       & \textbf{73.8}                         & \textbf{7.0}                           & \multicolumn{1}{c|}{\textbf{55.6}}      & \textbf{70.1}                       & \textbf{63.9}                         & 8.9                                    & \multicolumn{1}{c|}{\textbf{49.3}}    & \textbf{76.3}                       & \textbf{49.6}                         & \textbf{10.1}                          & \textbf{48.3}    \\ \hline 
\end{tabular}
\vspace*{-3mm}
\caption{Performance of VLN-GOAT under various token pruning strategies. Metrics include navigation success rate (``Seen'',``Unseen''), average steps from both splits, and average FLOPS\% relative to the base model. 
 FastV$^1$\cite{chen2025image} (textual \& visual), Cascade pruning$^2$ (textual \& visual) \cite{goyal2020power,wang2021spatten}, 
Cascade pruning and Token Merging \cite{bolya2022token} (Cascade + ToMe$^3$, visual)
are pruning techniques that are compared to NAP. NAP consistently yields better FLOPS reduction while maintaining comparable or higher SRs. 
}
\label{tab:GOAT-performance}
\end{table*}

\noindent
\textbf{Datasets.} We evaluated NAP on R2R~\cite{anderson2018vision}, RxR-English~\cite{ku2020room}, and REVERIE~\cite{qi2020reverie}. For each dataset, token pruning was tested on two splits, the validation “seen” split and validation “unseen” split. 

\noindent
\textbf{Evaluation Metrics.} Our primary {\em efficacy} metric is navigation Success Rate (SR), which measures the agent's ability to navigate correctly.
Additional efficacy metrics are reported in our result tables.
For {\em efficiency}, we use a simplified version of  
FLOPS-based formula from \citet{wang2024magic}:
$$G_\text{total} \!= \! G_{lan}(I) + D \!\cdot \!(G_{vis}(P) + G_{cm}(I,P,V)) \!+ \!c,$$ 
where $D$ is the number of decision steps
, $G_{lan}$ the Giga-FLOPS (GFLOPS) of the language module getting instruction features, $G_{vis}$ the GFLOPS of the visual module processing view features, $G_{cm}$ the GFLOPS of the cross-modal module predicting actions from all input features, and $c$ the cost from other sources, if any.
To 
convey efficiency improvements, we report the ratio of GFLOPS after pruning to GFLOPS before pruning [in \%]:
\begin{equation}
\label{eq:GFLOPS}
\text{FLOPS\%} = G_\text{pruned}/G_\text{original}. 
\vspace*{-2mm}
\end{equation}
We compute FLOPS using the Python toolkit \texttt{thop}.

\begin{table}[t]
\centering
\setlength{\tabcolsep}{2.5pt}
\begin{tabular}{ccc|ccc|c}
\hline
BGP                   & BTP                   & VPP                   & \multicolumn{1}{c}{SR$\uparrow$} & \multicolumn{1}{c}{SPL$\uparrow$} & \multicolumn{1}{c|}{Steps$\downarrow$} & \multicolumn{1}{c}{FLOPS\%$\downarrow$} \\ \hline
\xmark & \xmark & \xmark & 69.6                             & 61.8                              & 8.2                                    & 100.0                                  \\
\cmark & \xmark & \xmark & 65.5                             & 55.0                              & 9.1                                    & 76.6                                   \\
\xmark & \cmark & \xmark & 68.2                             & 60.6                              & 8.2                                    & 96.0                                   \\
\xmark & \xmark & \cmark & 68.2                             & 60.0                              & 8.4                                    & 81.9                                   \\
\cmark & \cmark & \xmark & 64.9                             & 55.3                              & 8.9                                    & 69.2                                   \\
\cmark & \xmark & \cmark & 65.4                             & 54.3                              & 9.3                                    & 56.5                                   \\
\xmark & \cmark & \cmark & 66.6                             & 58.8                              & 8.4                                    & 77.5                                   \\
\cmark & \cmark & \cmark & 63.9                             & 54.3                              & 9.0                                    & 49.3                                   \\ \hline
\end{tabular}
\vspace*{-3mm}
\caption{Ablation of BGP, BTP, and VPP on RxR-English Unseen. Each component lowers navigation cost while maintaining most SR and SPL.}
\label{tab:component_ablation}
\end{table}

\noindent
\textbf{Tested VLN Models.} We evaluated our token pruning strategies on three VLN models: HAMT~\cite{chen2021history}, DUET~\cite{chen2022think}, and GOAT~\cite{wang2024vision},
using 
the pre-tuned parameters provided by the respective authors. We extensively tested NAP with 
VLN-GOAT (Table~\ref{tab:GOAT-performance}) due to its outstanding performance across datasets. VLN-GOAT comprises six transformer layers in the language module, two in the view module, and three in the cross-modal module.

\noindent
\textbf{Tested Pruning Strategies.}
We are the first to apply \textit{multimodal} token pruning to VLN. As baselines, we evaluate general strategies applicable to VLN inputs: random pruning, cascade pruning~\cite{goyal2020power,wang2021spatten}, and FastV~\cite{chen2025image}. We also include Token Merging (ToMe)~\cite{bolya2022token} for view pruning. Unless specified, we only prune background views for visual input, due to significant SR drops (see ``Visual Input Pruning'').
\subsection{Results}
Table~\ref{tab:GOAT-performance} reports that NAP consistently achieves greater FLOPS reductions than baseline methods for VLN-GOAT. For example, at the 50\% pruning setting, NAP lowers FLOPS compared to prior work by  by 7 (R2R), 2.2 (RxR-English), and 3.1\% (REVERIE) points. 
At the same time, NAP also yields up to 2.5\% better navigation success rates.
 These advantages are retained even with larger token budget. For instance, in R2R at a 60\% pruning rate, NAP achieves nearly the same SR as Cascade + ToMe (a 0.4~pp difference) while reducing FLOPS by an additional 14~pp (from 74.4\% to 60.0\%). This improvement aligns with the trade-off curves shown in Fig.~\ref{fig:figure_1_1}.

In all R2R and REVERIE settings, NAP completes navigation with the fewest steps. Notably, under a 50\% token budget in R2R, NAP reduces the average number of steps from 7.3 to 7.0, while other strategies increase it. 

In Appendix~\ref{app:HAMT-DUET-tables} we also validated our method on HAMT~\cite{chen2021history} and DUET~\cite{chen2022think} using the R2R and REVERIE datasets, compared to FastV~\cite{chen2025image}. 
In DUET, our method outperforms FastV by 5pp to 10pp in SR, SPL, RGS, and RGSPL with similar FLOPS. In HAMT, where BTP cannot be applied, our approach still shows superior SR and SPL on R2R and outperforms FastV in both efficacy and efficiency on REVERIE.




\subsection{Model Analysis} 
Table~\ref{tab:component_ablation} assesses the effectiveness of BGP, BTP, and VPP under a 50\% token budget. We find
BGP achieves the largest FLOPS reduction (23.4pp), highlighting the high prunability of background views.  BTP contributes an additional 7~pp FLOPS reduction and shortens navigation by 0.3 steps on average (more than 1 on R2R), while VPP reduces FLOPS by 20pp.
We find that FLOPS improvement in VPP is closely related to the instruction length (see
Appendix~\ref{app:instr_lens}).

\begin{figure}[t]
\centering
\hspace*{-5mm}
\includegraphics[width=0.88\columnwidth]{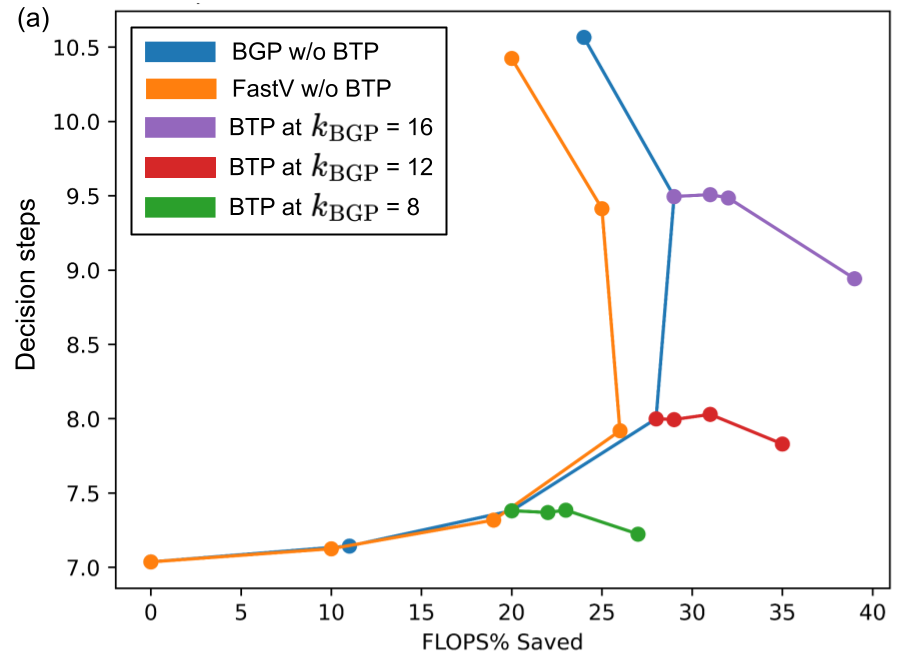}

\includegraphics[width=0.95\columnwidth]{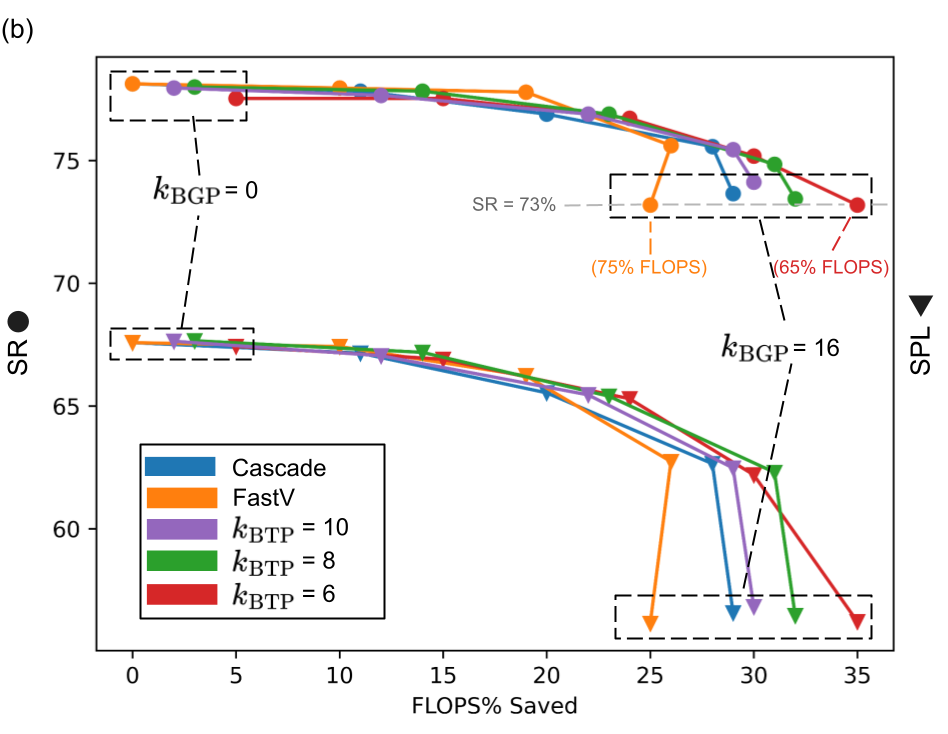}
\vspace*{-4mm}
\caption{ 
(a) Averaged decision steps for navigation applying BGP, FastV, and BTP. (b) SR and FLOPS saved of different pruning rates for Cascade pruning, FastV, and BGP with various BTP settings in R2R. BTP saves additional FLOPS by shortening the navigation length while keeping SR almost unaffected.}
\label{fig:BTP_ablation}
\end{figure}

\noindent
\textbf{Visual Input Pruning (BGP and BTP).} 
Table~\ref{tab:bgp_ablation} compares SR–FLOPS trade-off between BGP and full view pruning with 
with FastV~\cite{chen2025image} across 5 pruning rates. Pruning action views severely degrades efficacy (0\% SR at high rates) and can increase FLOPS 
(108\% at 20\%). This highlights the importance of preserving action views, a core aspect of BGP.
However, pruning background views increases decision steps (Fig.~\ref{fig:BTP_ablation}(a), from 7.0 to 10.5).  BTP reverses this trend by keeping just 6–10 unvisited nodes per step. As shown in Fig.~\ref{fig:BTP_ablation}(b), doing so saves 10 pp more FLOPS than FastV (pruning backgrounds), with minimal SR loss.  Appendix~\ref{app:btp_alone_generalization} reports that BTP also improves performance on our other datasets.



\begin{figure}[t]
\centering
\includegraphics[width=0.9\columnwidth]{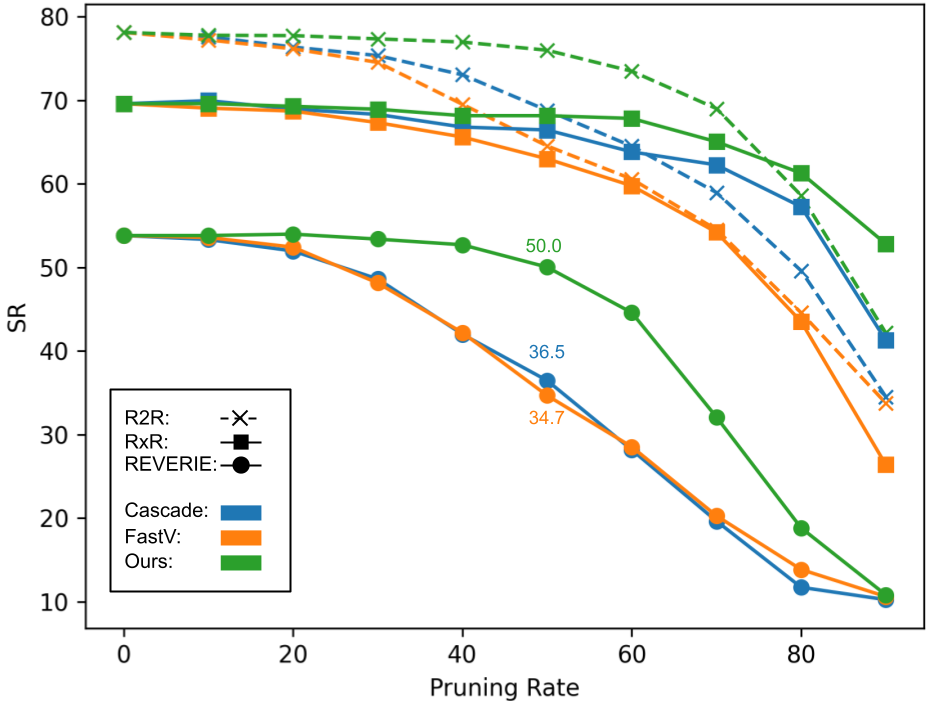}
\vspace*{-3mm}
\caption{Success rates of VLN applying VPP, Cascade Pruning, and FastV on instruction input. VPP preserves noticeably more SR with the same pruning rate than Cascade Pruning and FastV.}
\label{fig:ablation_vpp}
\end{figure}

\noindent\textbf{Textual Input (VPP)}. VPP prunes the textual tokens in the instruction $I$. We compare VPP with Cascade Pruning and FastV for instruction token removal on different datasets (Fig.~\ref{fig:ablation_vpp}). VPP (green) consistently outperforms the other two strategies across most pruning rates, with improvements >10~pp in the 40–60\% range. 

Table~\ref{tab:vpp_cross_model} evaluates whether a vocabulary constructed from one VLN dataset can be reused on other datasets. We can see from the result that cross-dataset vocabularies still effectively mitigate performance loss.  For example, a vocabulary built from RxR achieves 75.5\% SR under 50\% pruning, close to the 76.0\% SR that is from the R2R dataset. This suggests that navigation-relevant words are largely shared across datasets.

\begin{table}[t]
\setlength{\tabcolsep}{0.5pt}

\hspace*{-3mm}
\centering
\begin{tabular}{c|ccc|ccc}
\hline
\multirow{2}{*}{\begin{tabular}[c]{@{}c@{}}Pruning\\ Rate\%\end{tabular}} & \multicolumn{3}{c|}{Full Views (FastV)}                        & \multicolumn{3}{c}{BGP}                               \\ \cline{2-7} 
                                                                          & SR$\uparrow$ & Steps$\downarrow$ & FLOPS\%$\downarrow$ & SR$\uparrow$ & Steps$\downarrow$ & FLOPS\%$\downarrow$ \\ \hline
0                                                                         & 78           & 7                 & 100                & 78           & 7                 & 100                \\
20                                                                        & 51           & 11                & 108                & 77           & 7                 & 80                 \\
40                                                                        & 3            & 2                 & 20                 & 74           & 9                 & 71                 \\
60                                                                        & 0            & 0                 & 10                 & 69           & 11                & 76                 \\
80                                                                        & 0            & 0                 & 9                  & 69           & 11                & 73                 \\ \hline
\end{tabular}
\vspace{-3mm}
\caption{Comparison of BGP and pruning full views including action views across different pruning rates on the R2R dataset. Pruning action views leads to substantial drops in SR or longer, less efficient navigation paths, indicating the importance of preserving action views.}
\label{tab:bgp_ablation}
\end{table}

\begin{table}[t]

\small 
\centering
\begin{tabular}{|c|ccc|}
\hline
\backslashbox{Target}{Source}                                        & R2R                                                    & RxR                                                    & REVERIE                                                \\ \hline
\begin{tabular}[c]{@{}c@{}}R2R\\ (68.8)\end{tabular}     & \begin{tabular}[c]{@{}c@{}}+7.2\\ (76.0)\end{tabular}  & \begin{tabular}[c]{@{}c@{}}+6.7\\ (75.5)\end{tabular}  & \begin{tabular}[c]{@{}c@{}}+5.2\\ (74.0)\end{tabular}  \\ \hline
\begin{tabular}[c]{@{}c@{}}RxR\\ (66.4)\end{tabular}     & \begin{tabular}[c]{@{}c@{}}+1.6\\ (68.0)\end{tabular}  & \begin{tabular}[c]{@{}c@{}}+1.8\\ (68.2)\end{tabular}  & \begin{tabular}[c]{@{}c@{}}+1.8\\ (68.2)\end{tabular}  \\ \hline
\begin{tabular}[c]{@{}c@{}}REVERIE\\ (36.5)\end{tabular} & \begin{tabular}[c]{@{}c@{}}+13.3\\ (49.8)\end{tabular} & \begin{tabular}[c]{@{}c@{}}+11.4\\ (47.9)\end{tabular} & \begin{tabular}[c]{@{}c@{}}+13.5\\ (50.0)\end{tabular} \\ \hline
\end{tabular}
\caption{Navigation success rates (in parentheses) using vocabularies of irrelevance built from source datasets (columns) and applied to target datasets (rows). The leftmost column (e.g., 68.8) shows success rates under a 50\% instruction pruning rate with Cascade Pruning as a reference. Performance remains consistent across vocabularies from different sources, suggesting that irrelevant vocabulary is largely dataset-independent.} 
\label{tab:vpp_cross_model}
\end{table}

\noindent
\textbf{Continuous Environment.} While our navigation so far is based on a node-to-node discrete setting, NAP also applies to continuous environments through a waypoint predictor. The waypoint predictor is a commonly adopted module~\cite{krantz2021waypoint,hong2022bridging,an2024etpnav} that converts low-level actions (e.g., “go left”) into high-level node traversals (e.g., “go to viewpoint A”), enabling discrete VLN models to operate in continuous settings. NAP can therefore leverage the outputs of the waypoint predictor for input pruning in continuous environments.

To validate this, Table~\ref{tab:VLN-CE} compares NAP and Fast-V on VLN-CE-R2R~\cite{krantz2020beyond} under a 50\% visual+text token budget, integrating them into the ETPNav~\cite{an2024etpnav} model. Given that the waypoint predictor introduces an additional computational overhead of approximately 10~pp FLOPS, NAP reduces 37.6~pp FLOPS with an SR drop of 4.2~pp, which is 14.1 pp higher than Fast-V. Meanwhile, we observe that NAP leads to roughly one additional step per navigation compared to Fast-V, which diminishes the FLOP savings from pruning backtracking nodes. We will leave the investigation of this phenomenon for further FLOPS improvement as future work.

Table~\ref{tab:VLN-CE_ablation} also provides an ablation study of NAP components in the continuous setting. VPP is the most effective strategy in this case: 24.1~pp FLOPS reduced with only a 1.3~pp SR drop. Compared to the discrete setting, FLOP reductions shift: BGP contributes far less (3~pp), while BTP and VPP contribute substantially more (10~pp and 24~pp, respec). This difference arises because VLN models in the continuous setting use a smaller view encoder to enable faster image processing.

\begin{table}[]
\setlength{\tabcolsep}{1pt}
\begin{tabular}{l|ccc|ccc}
\hline
Strategy & $k_{BGP}$ & $k_{BTP}$ & VPP  & Steps$\downarrow$ & SR$\uparrow$   & FLOPS\%$\downarrow$ \\ \hline
NAP      & 3   & 4   & 40\% & 9.6   & 52.7 & 62.4 \\
FastV    & 3   & -   & 40\% & 8.3   & 38.6 & 63.9 \\ \hline
\end{tabular}
\caption{A comparison between NAP and Fast-V on VLN-CE-R2R dataset~\cite{krantz2020beyond} with a total token budget of 50\%. NAP preserves 14.1~pp more SR while achieving similar FLOPS deduction.}
\label{tab:VLN-CE}
\end{table}

\begin{table}[]
\centering
\setlength{\tabcolsep}{4pt}
\begin{tabular}{ccc|ccc}
\hline
BGP & BTP & VPP & \multicolumn{1}{c}{SR$\uparrow$} & \multicolumn{1}{c}{{\color[HTML]{333333} Steps$\downarrow$}} & \multicolumn{1}{c}{{\color[HTML]{333333} FLOPS\%$\downarrow$}} \\ \hline
\xmark               & \xmark               & \xmark                           & 56.9                         & {\color[HTML]{333333} 8.8}                                & {\color[HTML]{333333} 100.0}                             \\
\xmark               & \cmark            & \xmark                           & 55.6                         & {\color[HTML]{333333} 9.2}                                & {\color[HTML]{333333} 90.8}                              \\
\cmark            & \xmark               & \xmark                           & 56.2                       & {\color[HTML]{333333} 8.8}                                & {\color[HTML]{333333} 97.4}                              \\
\cmark            & \cmark            & \xmark                           & 54.0                         & {\color[HTML]{333333} 9.3}                                & {\color[HTML]{333333} 88.9}                              \\
\xmark               & \xmark               & \cmark                     & 55.6                         & {\color[HTML]{333333} 9.2}                                & {\color[HTML]{333333} 75.9}                              \\
\xmark               & \cmark            & \cmark                     & 53.0                         & {\color[HTML]{333333} 9.7}                                & {\color[HTML]{333333} 65.3}                              \\
\cmark            & \xmark               & \cmark                     & 56.1                         & {\color[HTML]{333333} 9.2}                                & {\color[HTML]{333333} 73.8}                              \\
\cmark            & \cmark            & \cmark                     & 52.7                         & {\color[HTML]{333333} 9.6}                                & {\color[HTML]{333333} 62.4}                              \\ \hline
\end{tabular}
\caption{Ablation study of NAP components on the VLN-CE dataset. The pruning parameters are the same as Table~\ref{tab:VLN-CE}. The result shows that all components reduce FLOPs, with BTP and VPP proving substantially more effective than BGP (9.2~pp and 24.1~pp vs. 2.6~pp, rows 2, 5, and 3).}
\label{tab:VLN-CE_ablation}
\end{table}


\begin{figure}[t]
\centering
\includegraphics[width=\columnwidth]{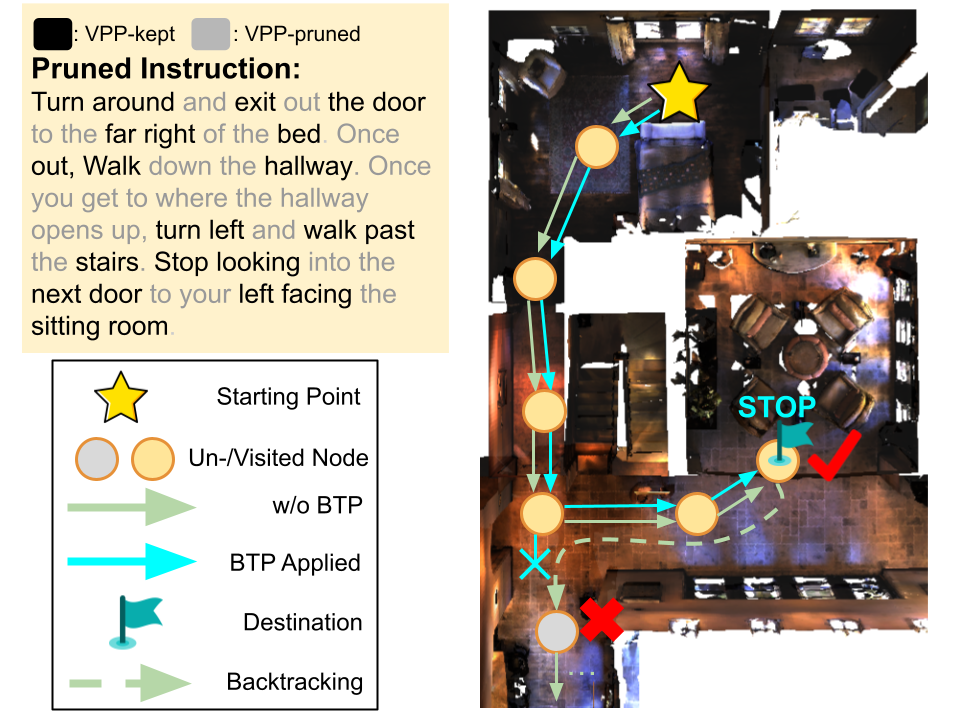}
\vspace*{-8mm}
\caption{An example of navigation with NAP. BTP prevents a long and failed navigation path by pruning unvisited nodes, while VPP preserves most key information and removes tokens unhelpful to navigation.}
\label{fig:btp_qual}
\end{figure}

\noindent
\textbf{Illustrative Result.}
The impact of NAP is illustrated in a navigation example in Fig.~\ref{fig:btp_qual}.
VPP preserves key instruction words, enabling the agent to 
complete the route. Meanwhile, BTP ``forces'' the agent to stop at the correct destination by removing potential backtracking nodes; without it, unpruned unvisited nodes would cause backtracking, resulting in longer paths and navigation failure.

\section{Conclusions}
Our work introduces NAP, a navigation-specific token pruning framework that integrates three strategies: BGP, BTP, and VPP. Compared to NAP, general pruning methods could cause longer navigation paths due to the view removal. Also, as methods designed for tasks other than navigation, they also fail to reliably preserve navigation-critical instruction tokens. NAP addresses these issues by tailoring pruning to the navigation task, reducing the input size of VLN models across all modalities: text, visual views, and history nodes. Specifically, BGP prunes irrelevant background views while preserving action views; BTP removes redundant backtracking nodes, shortening both history inputs and navigation paths; and VPP employs an LLM-guided vocabulary to distinguish and retain navigation-relevant tokens, enabling more aggressive yet precise pruning. Experiments on the R2R, RxR-English, and REVERIE datasets show that NAP achieves deeper token reduction while maintaining higher success rates in VLN tasks. Our model- and dataset-agnostic framework further reduces computational cost while improving navigation efficiency, enabling the deployment of VLN models in resource-constrained environments.


\section{Limitations}
A majority of generic token pruning strategies, including our BGP and BTP, assume that attention scores accurately reflect the true importance of token features for navigation success. However, this assumption is inherently flawed. While VPP partly mitigates this issue for instruction pruning, an analogous solution for view pruning would incur expensive costs, such as those associated with gradient- or perturbation-based saliency. Moreover, as shown in Table~\ref{tab:GOAT-performance}, random token pruning performs nearly as well as attention-based methods—except for our approach. This
suggests that 
many 
VLN tokens may 
contribute similarly to navigation, yielding comparable efficiency gains and success rate losses regardless of the pruning method. We hope our work serves as a starting point for understanding token importance in VLN for both text and visuals, and inspires 
further
development of 
effective, cost-efficient indicators for VLN token pruning.

\bibliography{references/custom}

\clearpage




\appendix

\section{VLN Datasets Statistics}
The statistic of VLN dataset statistic is given in Table~\ref{tab:stat}. The tokens numbers are averaged over steps of all navigation tasks. The statistics show that R2R tasks are easier with shorter paths and instructions. On the other hand, RxR-English tasks contain longer instructions and paths.

\begin{table}[t]
\small
\setlength{\tabcolsep}{1pt}
\begin{tabular}{|c|cc|cc|cc|}
\hline
\multirow{2}{*}{Dataset} & \multicolumn{2}{c|}{R2R} & \multicolumn{2}{c|}{RxR-english} & \multicolumn{2}{c|}{REVERIE} \\ \cline{2-7} 
                         & Seen       & Unseen      & Seen           & Unseen          & Seen         & Unseen        \\ \hline
Size                     & 1020       & 2349        & 2939           & 4551            & 1423         & 3521          \\ \hline
History nodes size       & \multicolumn{2}{c|}{14}  & \multicolumn{2}{c|}{19}          & \multicolumn{2}{c|}{16}      \\ \hline
View tokens              & \multicolumn{2}{c|}{38}  & \multicolumn{2}{c|}{38}          & \multicolumn{2}{c|}{42}      \\ \hline
Instruction tokens       & \multicolumn{2}{c|}{32}  & \multicolumn{2}{c|}{127}         & \multicolumn{2}{c|}{20}      \\ \hline
\end{tabular}
\caption{Statistics of R2R, RxR-English, and REVERIE regarding token pruning. The numbers are averaged over all navigation tasks and steps. R2R is easier compared to RxR-English and REVERIE given its path and instruction lengths.}
\label{tab:stat}
\end{table}


\section{NAP Performance of HAMT and DUET Models on R2R and REVERIE Datasets}

\label{app:HAMT-DUET-tables}

\begin{table}[t]
\small
\setlength{\tabcolsep}{3pt}
\begin{tabular}{l|c|ccccc}
\hline
\multirow{2}{*}{Method} & \multirow{2}{*}{Tokens} & \multicolumn{5}{c}{Validation Unseen} \\ \cline{3-7}

                                                                        &                                                                             & SR$\uparrow$  & SPL$\uparrow$  & RGS$\uparrow$  & RGSPL$\uparrow$  & FLOPS\%$\downarrow$ \\ \hline
HAMT                                                                    & -                                                                         & 32   &   29   &  19    &    17    &    100      \\ \hline \hline
\multirow{3}{*}{\begin{tabular}[c]{@{}l@{}}HAMT\\ (FastV)\end{tabular}} & 70\%                                                                          &  \textbf{32}   &  29    &  18    &   16     &  73        \\
                                                                        & 60\%                                                                          &  30   &  27    &  16    &   14     &    68      \\
                                                                        & 50\%                                                                          &  27   &  24    &   14   &   12     &   65       \\ \hline
\multirow{3}{*}{\begin{tabular}[c]{@{}l@{}}HAMT\\ (NAP$^{-}$)\end{tabular}}  & 70\%                                                                          &  31   &  29    &  \textbf{19}    &    \textbf{17}    &    \textbf{69}     \\
                                                                        & 60\%                                                                          &  \textbf{31}   &  \textbf{28}    &  \textbf{18}    &   \textbf{17}     &   \textbf{64}       \\
                                                                        & 50\%                                                                          &  \textbf{29}   &  \textbf{26}    &  \textbf{17}    &  \textbf{15}     &   \textbf{59}       \\ \hline
                                                                        \hline
DUET                                                                    & -                                                                         & 47  & 34   & 32   & 23     & 100      \\ \hline
\multirow{3}{*}{\begin{tabular}[c]{@{}l@{}}DUET\\ (FastV)\end{tabular}} & 70\%                                                                          & 39  & 28   & 23   & 16     & 82       \\
                                                                        & 60\%                                                                          & 34  & 24   & 19   & 13     & 73       \\
                                                                        & 50\%                                                                          & 30  & 20   & 16   & 11     & 64       \\ \hline
\multirow{3}{*}{\begin{tabular}[c]{@{}l@{}}DUET\\ (NAP)\end{tabular}}  & 70\%                                                                          & \textbf{47}  & \textbf{33}   & \textbf{31}   & \textbf{22}     & \textbf{80}       \\
                                                                        & 60\%                                                                          & \textbf{45}  & \textbf{32}   & \textbf{31}   & \textbf{22}     & \textbf{70}       \\
                                                                        & 50\%                                                                          & \textbf{43 } & \textbf{30}   & \textbf{28}   & \textbf{20}     & \textbf{62}       \\ \hline
\end{tabular}
\vspace*{-3mm}
\caption{Performance of pruning strategies on 
REVERIE data
NAP$^{-}$ indicates pruning without BTP.
Efficacy is measured in Success Rate (SR) Per Length (SPL),  Remote Grounding Success rate (RGS), and Remote Grounding Success rate Per Length (RGSPL). The results show that NAP is also effective for different navigation tasks such as remote object localization. 
}
\label{tab:reverie}
\end{table}

We also compared the NAP performance of HAMT and DUET models on the REVERIE (Table~\ref{tab:reverie}) and R2R (Table~\ref{tab:r2r}) datasets against FastV. In the HAMT setting, we find that NAP achieves comparable or superior results, even without BTP (since HAMT lacks a history mapping module). For example, on REVERIE (Table~\ref{tab:reverie}), NAP achieves greater FLOP reductions than FastV (6 pp to 4 pp) while maintaining similar or better performance in both navigation (SR, SPL) and object localization (RGS, RGSPL) across token budgets ranging from 70\% to 50\%. The advantage of NAP is even more pronounced for the DUET model: NAP consistently yields a better trade-off, with navigation performance (SR) surpassing FastV by 8–13~pp, and FLOP reductions exceeding FastV by 2–3~pp. Similar observation can be found on R2R dataset as well (Table~\ref{tab:r2r}), showing that NAP benefits different VLN models, even without a history mapping.

\begin{table}[t]
\small
\setlength{\tabcolsep}{1pt}
\begin{tabular}{l|c|ccc|ccc}
\hline
\multirow{2}{*}{Method}                                                 & \multirow{2}{*}{\begin{tabular}[c]{@{}c@{}}Retain\\ Rate\end{tabular}} & \multicolumn{3}{c|}{Validation Seen} & \multicolumn{3}{c}{Validation Unseen} \\ \cline{3-8} 
                                                                        &                                                                             & SR$\uparrow$       & SPL$\uparrow$       & FLOPS\%$\downarrow$      & SR$\uparrow$       & SPL$\uparrow$       & FLOPS\%$\downarrow$       \\ \hline \hline
HAMT                                                                    & -                                                                        &    70      &     67      &      100         &     63     &      58    &      100          \\ \hline
\multirow{3}{*}{\begin{tabular}[c]{@{}l@{}}HAMT\\ (FastV)\end{tabular}} & 0.7                                                                          &    67      &     64      &      80         &     \textbf{60}     &   \textbf{54}        &     82           \\
                                                                        & 0.6                                                                          &   63       &    60       &        72       &    56      &    50       &          75      \\
                                                                        & 0.5                                                                          &    58      &    55       &      \textbf{65}         &    50      &   45        &   68             \\ \hline
\multirow{3}{*}{\begin{tabular}[c]{@{}l@{}}HAMT\\ (NAP$^{-}$)\end{tabular}}  & 0.7                                                                          &    \textbf{71}      &     \textbf{66}      &        \textbf{79}       &      59    &    53       &     \textbf{ 81}          \\
                                                                        & 0.6                                                                          &   \textbf{69}       &      \textbf{65}     &    72           &   56      &    \textbf{51}      &   \textbf{74}            \\
                                                                        & 0.5                                                                          &    \textbf{63}      &      \textbf{60}     &    66           &     \textbf{52}     &     \textbf{46}      &  68              \\ \hline
                                                                        \hline
DUET                                                                    & -                                                                         & 79       & 73        & 100           & 72       & 60        & 100            \\ \hline 
\multirow{3}{*}{\begin{tabular}[c]{@{}l@{}}DUET\\ (FastV)\end{tabular}} & 0.7                                                                          & 68       & 62        & 73            & 60       & 50        & \textbf{73}             \\
                                                                        & 0.6                                                                          & 62       & 56        & 66            & 54       & 44        & \textbf{65}             \\
                                                                        & 0.5                                                                          & 59       & 52        & \textbf{56}            & 50       & 40        & \textbf{55}             \\ \hline
\multirow{3}{*}{\begin{tabular}[c]{@{}l@{}}DUET\\ (NAP)\end{tabular}}  & 0.7                                                                          & \textbf{77}       & \textbf{65}        & \textbf{71}            & \textbf{68}       & \textbf{57}        & 75             \\
                                                                        & 0.6                                                                          & \textbf{75}       & \textbf{69}        & \textbf{64}            & \textbf{66}       & \textbf{55}        & 68             \\
                                                                        & 0.5                                                                          & \textbf{72}       & \textbf{71}        & 57            & \textbf{63}       & \textbf{52}        & 59             \\ \hline
\end{tabular}
\vspace*{-3mm}
\caption{Model performance on the R2R dataset with NAP and FastV. NAP$^{-}$ indicates pruning without BTP. The result shows that NAP can be adapted to different models while still outperforming FastV.}
\label{tab:r2r}
\end{table}

\section{BTP combined with different view pruning strategies}
\label{app:btp_alone_generalization}

\begin{figure}[h]
\centering
\hspace*{-5mm}
\includegraphics[width=0.82\columnwidth]{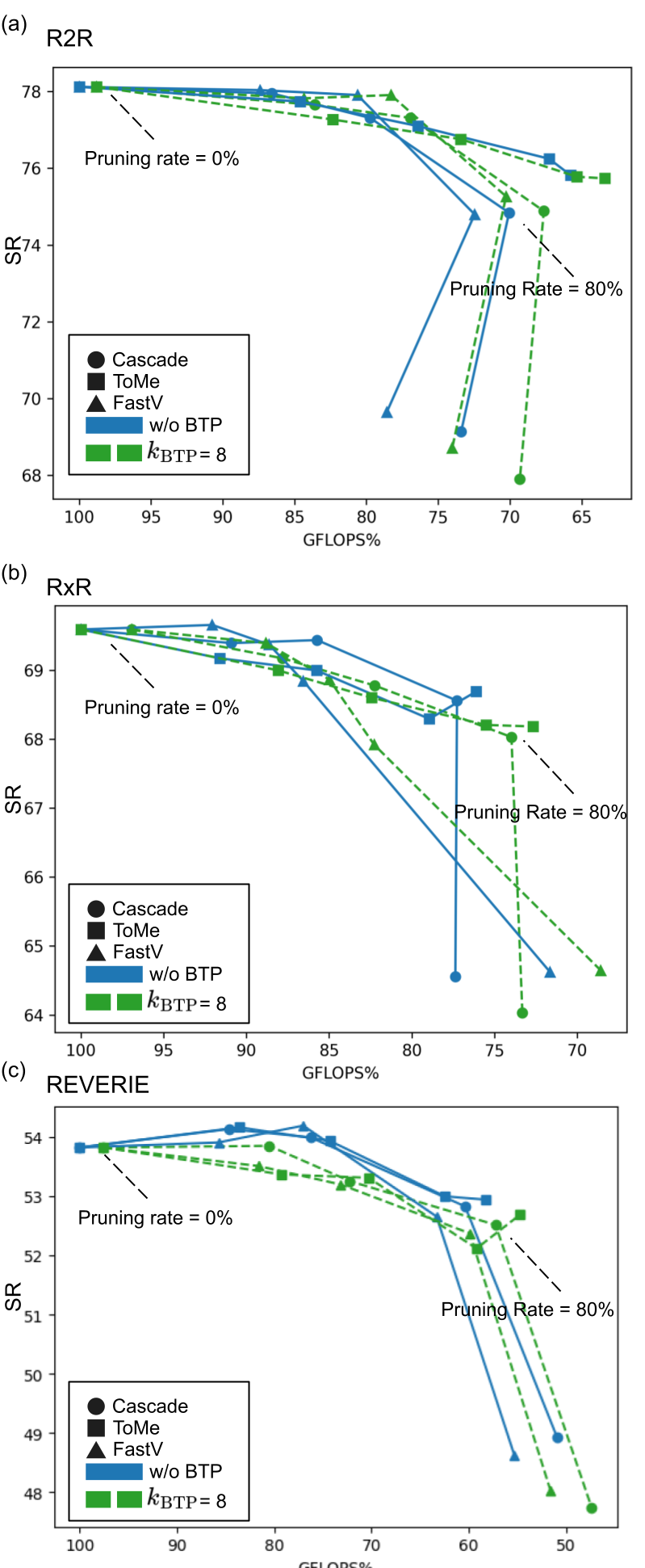}
\caption{SR and FLOPS curves for different view pruning rates with and without BTP. Performance is evaluated on the R2R, RxR and REVERIE datasets.}
\label{fig:btp_ablation_app}
\end{figure}

We evaluated the FLOPS improvements when applying BTP respectively with cascade token pruning, FastV, and Token Merging (ToMe) across the R2R, RxR, and REVERIE datasets (see Fig.~\ref{fig:btp_ablation_app}). Our findings show that incorporating BTP reduces FLOPS by approximately 5 percentage points at high BGP rates (e.g., 80\%), while resulting in a success rate drop of less than 0.5 percentage points.

\section{Instruction Length Influence on Efficiency Improvement}

\label{app:instr_lens}
\begin{figure}[h]
\centering
\hspace*{-5mm}
\includegraphics[width=\columnwidth]{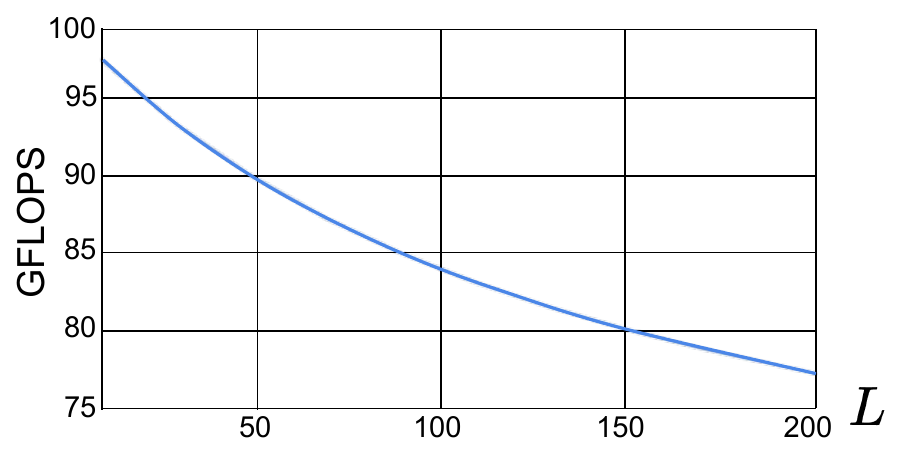}
\caption{FLOPS curve over different instruction lengths with 50\% token pruning rate. The curve shows that FLOPS reduced from textual input is heavily dependent on the instruction length, varying from 10~pp given 50 tokens to 20~pp given 150 tokens.}
\label{fig:instr_lens}
\end{figure}

VPP's efficiency gains are closely related to instruction length. Under the default settings for R2R, RxR, and REVERIE (36 views, 14 history nodes, and a 7-step navigation), we evaluated the efficiency improvement from a 50\% instruction pruning rate across different instruction lengths (see Fig.~\ref{fig:instr_lens}). The results show that a 50-token instruction achieves a 10 percentage point FLOPS reduction, which increases to 20 percentage points for a 150-token instruction.

\begin{table}[t]
\centering
\setlength{\tabcolsep}{1.5pt}
\begin{tabular}{l|ccc}
\hline
Method & \multicolumn{1}{l}{Steps$\downarrow$} & \multicolumn{1}{l}{SR$\uparrow$} & \multicolumn{1}{l}{FLOPS\%$\downarrow$} \\ \hline
No pruning      & 7                                          & 78.1                            & 100.00\%                             \\
SAS             & 6.8                                        & 76.8                            & 96.45\%                              \\
NAP (70\%)      & 7.2                                        & 74.5                            & 69.04\%                              \\
NAP+SAS (70\%)  & 7.5                                        & 72.3                            & 71.22\%                              \\
NAP             & 7.8                                        & 75.2                            & 60.92\%                              \\
NAP+SAS (60\%)  & 7.9                                        & 72.2                            & 61.53\%                              \\
NAP             & 7.9                                        & 73.8                            & 57.03\%                              \\
NAP+SAS (50\%)  & 8                                          & 71.2                            & 57.72\%                              \\ \hline

\end{tabular}
\caption{A comparison with SAS and NAP under different pruning rates. NAP preserves higher SR than SAS, while FLOPS reduction is similar for both.}
\label{tab:SAS}
\end{table}

\begin{table}[h!]
\centering
\begin{tabular}{l|cc}
\hline
\multicolumn{1}{c|}{Method} & \begin{tabular}[c]{@{}c@{}}Cross Attention \\ FLOPS\end{tabular} & \begin{tabular}[c]{@{}c@{}}Transformer Layer\\ FLOPS\end{tabular} \\ \hline
SAS                         & $4.4 * 10^{-4}$                   & 0.114                    \\
No SAS                      & $4.0 * 10^{-3}$                  & 0.117                    \\ \hline
\end{tabular}

\caption{Analysis of FLOPS reduced by applying SAS on a single transformer layer. SAS only saves very little (0.003) FLOPS in a VLN transformer layer.}
\label{tab:SAS_transformer}
\end{table}

\begin{table*}[t]
\small
\centering
\begin{tabular}{|c|cl|}
\hline
Prompt  & \multicolumn{2}{l|}{\begin{tabular}[c]{@{}l@{}}Given the following set of words: ``walk'', ``down'', ..., ``landing'', \\ can you point out which of them are irrelevant to the following types of information: \\ 1. A direction to go; 2. Describing the environment; 3. Object(s) in the indoor/outdoor environments. \\ Please don't change the word in the quotation mark and explain why. \\ Please answer in the following: format: ``\{word\} : relevant/irrelevant \{explanation\}''\end{tabular}} \\ \hline
Word    & \multicolumn{1}{c|}{Relevance}                                                                                                                                                                                                & \multicolumn{1}{c|}{Explanation}                                                                                                                                                                                                                               \\ \hline
walk    & \multicolumn{1}{c|}{relevant}                                                                                                                                                                                                 & direction to go                                                                                                                                                                                                                                                \\ \hline
down    & \multicolumn{1}{c|}{relevant}                                                                                                                                                                                                 & direction to go                                                                                                                                                                                                                                                \\ \hline
one     & \multicolumn{1}{c|}{irrelevant}                                                                                                                                                                                               & not related to direction or environment description                                                                                                                                                                                                            \\ \hline
flight  & \multicolumn{1}{c|}{irrelevant}                                                                                                                                                                                               & not related to indoor environment navigation                                                                                                                                                                                                                   \\ \hline
of      & \multicolumn{1}{c|}{irrelevant}                                                                                                                                                                                               & preposition, not related to direction or environment description                                                                                                                                                                                               \\ \hline
stairs  & \multicolumn{1}{c|}{relevant}                                                                                                                                                                                                 & object in the environment                                                                                                                                                                                                                                      \\ \hline
and     & \multicolumn{1}{c|}{irrelevant}                                                                                                                                                                                               & conjunction, not related to direction or environment description                                                                                                                                                                                               \\ \hline
stop    & \multicolumn{1}{c|}{relevant}                                                                                                                                                                                                 & direction to go                                                                                                                                                                                                                                                \\ \hline
the     & \multicolumn{1}{c|}{irrelevant}                                                                                                                                                                                               & article, not related to direction or environment description                                                                                                                                                                                                   \\ \hline
landing & \multicolumn{1}{c|}{relevant}                                                                                                                                                                                                 & object in the environment                                                                                                                                                                                                                                      \\ \hline
\end{tabular}
\caption{Our prompt to Llama 3, and examples of words in the vocabulary of irrelevance, explained by the LLM why they are relevant/irrelevant.}
\label{tab:vocabulary}
\end{table*}

\section{Comparison between Pruning Tokens and Skipping Attention Calculation Only}

We observe that token-related computational cost arises from two sources: the linear projection ($O(ld^2)$) and the attention computation ($O(l^2d)$), where $l$ denotes the token count and $d$ the hidden feature size. Token pruning reduces both by decreasing $l$. As a lighter alternative, one can reduce FLOPs by shrinking $l$ only during the attention computation while retaining the full token set in the projection layers. We term this approach Selective Attention Sum (SAS). SAS restricts attention from background (non-explorable) views to action (explorable) views, thereby skipping unnecessary computations.

We evaluate SAS in combination with NAP on the R2R unseen split (Table~\ref{tab:SAS}). SAS introduces a 1–3~pp drop in success rate (SR) and slightly increases navigation steps under tight token budgets, which offsets its marginal FLOPS savings.

To further analyze the impact of saving computation during attention computation, we compared the computation of a transformer layer with and without SAS in terms of GFLOPS, shown in Table~\ref{tab:SAS_transformer}. While SAS reduces attention score computations (from 
 $4.0 * 10^{-3}$  to $4.4 * 10^{-4}$ 
 GFLOPS across 36 views), the majority of computation in a transformer layer comes from linear projections. As a result, applying SAS only reduces the FLOPS of a transformer layer from 0.117 to 0.114, which is just a 2.5\% gain.

 We attribute the limited efficiency gain of SAS to the relatively small number of view tokens ($l=36$
) compared to the hidden size ($d=768$
), which makes the cost of linear projections ($O(ld^2)$
) dominant over attention computation ($O(l^2d)$
). In contrast, token pruning reduces both attention and projection costs, making it more effective than SAS for improving efficiency.


\section{Vocabulary Example}
\label{app:vocab_exam}
We provide an example of how we prompt the LLM to construct the vocabulary of irrelevance, with explanations from Llama 3 why such tokens are classified to be relevant or irrelevant (see Table~\ref{tab:vocabulary}).

\end{document}